\documentclass[runningheads]{llncs}
\usepackage{graphicx}

\usepackage{tikz}
\usepackage{comment}
\usepackage{amsmath,amssymb} 
\usepackage{color}
\usepackage{booktabs}
\usepackage{multirow}
\usepackage{bbding}
\usepackage{overpic}
\usepackage[breaklinks=true,bookmarks=false,colorlinks,citecolor=black,linkcolor=black]{hyperref}

\usepackage[accsupp]{axessibility}  

\def\eg{{\em e.g.}}

\def\etal{{\em et al.}}

\begin{document}
\pagestyle{headings}
\mainmatter
\def\ECCVSubNumber{4018}  

\title{High-Fidelity Image Inpainting with \\ GAN Inversion} 


\titlerunning{High-Fidelity Image Inpainting with GAN Inversion}
%
\author{Yongsheng Yu\inst{1,2} \and
Libo Zhang\inst{1,2,3}\thanks{Corresponding author (libo@iscas.ac.cn).} \and
Heng Fan\inst{4} \and
Tiejian Luo\inst{2}}

%
\authorrunning{Y. Yu, L. Zhang, H. Fan, T. Luo}
%
\institute{Institute of Software, Chinese Academy of Sciences \and
University of Chinese Academy of Sciences \and
Nanjing Institute of Software Technology \and
Department of Computer Science and Engineering, University of North Texas \\
\email{yuyongsheng19@mails.ucas.ac.cn}; \email{libo@iscas.ac.cn}; \email{heng.fan@unt.edu}; \email{tjluo@ucas.ac.cn}
}
\maketitle

\begin{abstract}
Image inpainting seeks a semantically consistent way to recover the corrupted image in the light of its unmasked content. Previous approaches usually reuse the well-trained GAN as effective prior to generate realistic patches for missing holes with GAN inversion. Nevertheless, the ignorance of a hard constraint in these algorithms may yield the gap between GAN inversion and image inpainting. Addressing this problem, in this paper, we devise a novel GAN inversion model for image inpainting, dubbed {\it InvertFill}, mainly consisting of an encoder with a pre-modulation module and a GAN generator with $\mathcal{F}\&\mathcal{W}^+$ latent space. Within the encoder, the pre-modulation network leverages multi-scale structures to encode more discriminative semantics into style vectors. In order to bridge the gap between GAN inversion and image inpainting, $\mathcal{F}\&\mathcal{W}^+$ latent space is proposed to eliminate glaring color discrepancy and semantic inconsistency. To reconstruct faithful and photorealistic images, a simple yet effective Soft-update Mean Latent module is designed to capture more diverse in-domain patterns that synthesize high-fidelity textures for large corruptions. Comprehensive experiments on four challenging datasets, including Places2, CelebA-HQ, MetFaces, and Scenery, demonstrate that our InvertFill outperforms the advanced approaches qualitatively and quantitatively and supports the completion of out-of-domain images well.


\keywords{Image Inpainting, GAN Inversion, $\mathcal{F}\&\mathcal{W}^+$ Latent Space}
\end{abstract}

\section{Introduction}

Image inpainting is an ill-posed problem that requires to recover the missing or corrupted content based on incomplete images with masks. It has been widely adopted for manipulating photographs, such as corrupted image repairing, unwanted object removal, or object position modification~\cite{criminisi2003object,shetty2018adversarial,song2019geometry}. 

The mainstream approaches~\cite{pathak2016context,Liu2018partial} often employ an encoder-decoder architecture in UNet style~\cite{ronneberger2015u} for image inpainting, and have demonstrated promising results in dealing with narrow holes or removing small objects. To apply to more complicated cases, later works have been focused on improving the performance with various discriminators~\cite{yu2019free,zeng2021aggregated,li2020recurrent}, contextual attention mechanisms~\cite{li2020recurrent,zeng2021aggregated,zeng2020high}, and auxiliary information~\cite{liu2020rethinking,zeng2021cr,nazeri2019edgeconnect,guo2021image}. Nevertheless, limited by their model capacity, it remains challenging for these UNet-like methods to fill large corruptions with visually realistic patches. 

Recently, generative adversarial network (GAN) models~\cite{karras2018progressive,karras2019style,karras2020analyzing} have been verified to successfully produce high-resolution photorealistic images. In these models, GAN inversion~\cite{zhu16generative,xia2021survey} plays an important role. Specifically, when simply fed with stochastic vectors of latent space, GAN is not applicable to any image-to-image translation. To handle this problem, GAN inversion method uses a pre-trained GAN as prior, and encodes the given images into stochastic vectors that represent the target images, resulting to high-fidelity translation results. Inspired by this, several approaches~\cite{gu2020mganprior,richardson2021encoding,cheng2021out} have made great efforts to introduced GAN inversion for image inpainting. Despite excellent performance, existing methods may suffer from following issues:

\begin{figure*}[t]
    \centering
	\includegraphics[width=0.61\columnwidth]{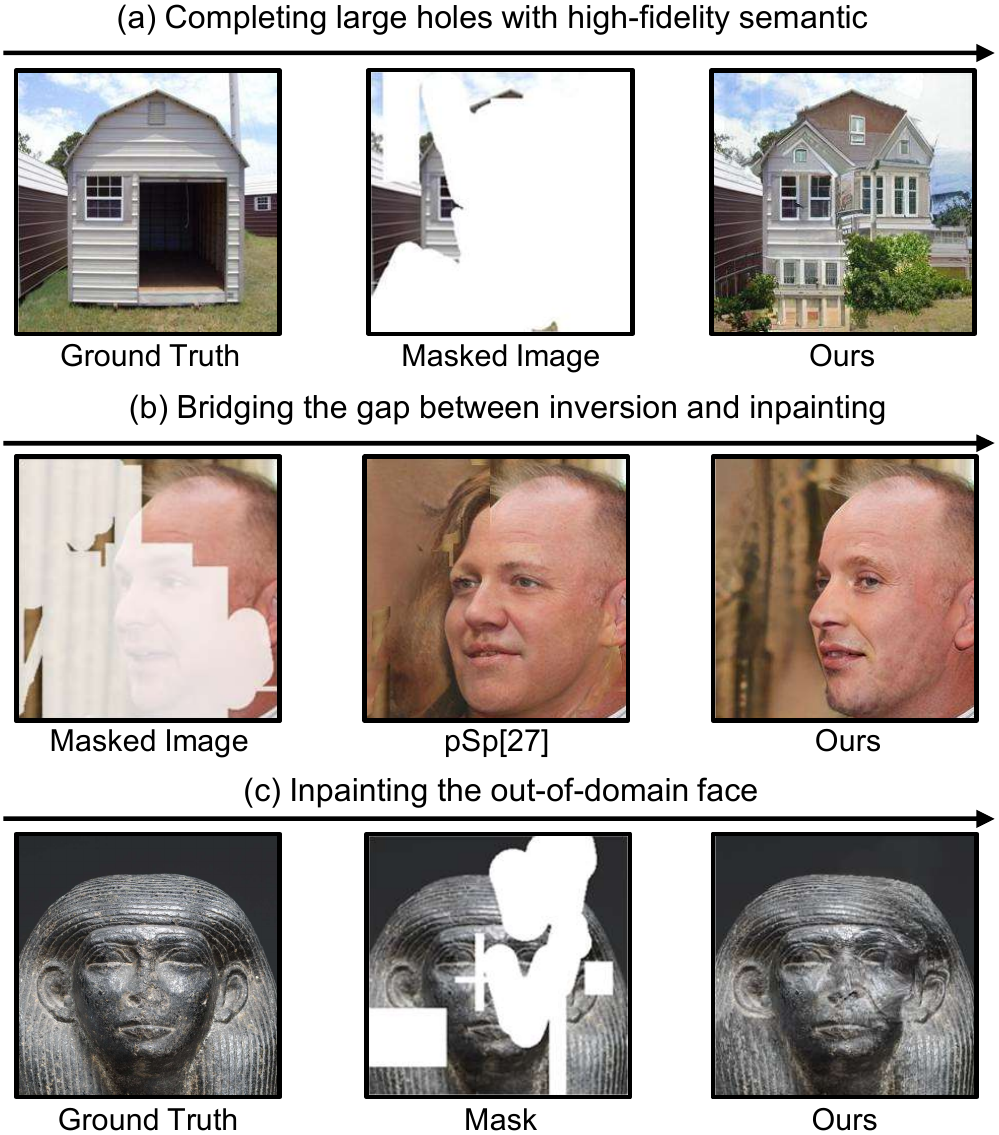}
	\caption{
Visual results of our contributions. Image (a) shows the high-fidelity inpainting results for large corruptions, image (b) exhibits the improvement of our method for the ``gapping'' issue over previous inversion-based inpainting method pSp~\cite{richardson2021encoding}, and image (c) demonstrates the semantically consistent results by our model for the out-of-domain masked image. {\it Best viewed in color for all figures throughout the paper.}
	}\label{fig:teaser}
\end{figure*}

\begin{itemize}
    \setlength{\itemsep}{3pt}
	\item \textit{\textbf{Distortion for extreme image inpainting.}} Due to large corruptions, 
	current methods (\eg,~\cite{li2020recurrent,zeng2020high,guo2021image}) may become degenerated because these models are {\it not} able to effectively extract correlation from inadequate knowledge in extremely degraded images. Such correlation information is crucial in eliminating the ambiguity of large continuous holes, especially where far from the boundary. 
	
	\item \textit{\textbf{Inconsistency caused by hard constraint.}} 
	Unlike in regular conditional translation (\eg, super-resolution~\cite{gu2020mganprior}, face editing~\cite{zhu2020domain} and label-to-image~\cite{richardson2021encoding}), image inpainting has a hard constraint that {\it the unmasked regions in the input and the output should be the same}. Current inversion-based algorithms~\cite{gu2020mganprior,richardson2021encoding,cheng2021out}, however, ignore this constraint, which results in color discrepancy and semantic inconsistency as displayed in Fig.~\ref{fig:teaser}(b) and may require additional post-processing such as image blending~\cite{cheng2021out}. We call this problem ``gapping'' in the following sections.

	
	\item \textit{\textbf{Robustness for out-of-domain inputs.}} 
	 In order to reconstruct faithful images, the key is to find an in-domain latent code that can align with the domain of a well-trained GAN model~\cite{zhu2020domain}. Unfortunately, the encoder fails to invert out-of-domain inputs to produce accurate results. For example, the pSp~\cite{richardson2021encoding} is hard to tackle the corrupted images with contents or masks from unseen domains, which is harmful to the applicability of GAN inversion.
	
\end{itemize}


To solve the above issues, we introduce a novel InvertFill network for image inpainting. It follows the encoder-based inversion fashion architecture~\cite{richardson2021encoding} that consists of an encoder and a GAN generator. We first develop a new latent space $\mathcal{F}\&\mathcal{W}^+$ (as explained later) that encodes the original images into style vector to enable the accessibility of the generator backbone to inputs, decreasing color discrepancy and semantic inconsistency. Besides, to make full use of the encoder, we present pre-modulation networks to amplify the reconstruction signals of the style vector based on the predicted multi-scale structures, further enhancing the discriminative semantic. Then, we propose a simple yet effective soft-update mean latent technique to sample a dynamic in-domain code for the generator. Compared to using a fixed code, our method is able to facilitate diverse downstream goals while reconstructing faithfully and photo-realistically, even in the task of unseen domain. To verify the superiority of our method, we conduct extensive experiments on four datasets, including CelebA-HQ~\cite{karras2018progressive}, Places2~\cite{zhou2017places}, MetFaces~\cite{DBLP:conf/nips/KarrasAHLLA20}, and Scenery~\cite{yang2019very}. The results demonstrate that our method achieves favorable performance, especially for images with large corruptions. Furthermore, our approach can handle images and masks from unseen domains by optimizing a lightweight encoder without retraining the GAN generator on a large-scale dataset. Fig.~\ref{fig:teaser} shows several visual results of our approach.

The contributions of our work are summarized as three-fold: {\bf (1)} We introduce a novel $\mathcal{F}\&\mathcal{W}^+$ latent space to resolve the problems of color discrepancy and semantic inconsistency and thus bridge the gap between image inpainting and GAN inversion. {\bf (2)} We propose (a) pre-modulation networks to encode more discriminative semantic from compact multi-scale structures and (b) soft-update mean latent to synthesize more semantically reasonable and visually realistic patches by leveraging diverse patterns. {\bf (3)} Extensive experiments on CelebA-HQ~\cite{karras2018progressive}, Places2~\cite{zhou2017places}, MetFaces~\cite{DBLP:conf/nips/KarrasAHLLA20}, and Scenery~\cite{yang2019very} show that the proposed approach outperforms current state-of-the-arts, evidencing its effectiveness.


\section{Related Work}

\noindent
{\bf Image Inpainting.}
Image inpainting could be treated as a conditional translation task with hard constraint. The seminal learning-based work by Pathak~\etal~\cite{pathak2016context} integrates UNet~\cite{ronneberger2015u} and GAN discriminator~\cite{goodfellow2014generative} for image inpainting, and subsequently derives many variants that effectively deal with narrow holes or remove small objects. More recently, several works have attempted to extending the idea in~\cite{pathak2016context} to more complicated cases. Roughly speaking, these methods can be categorized into three types. The first one is to explicitly dispose of invalid signals at masked regions~\cite{Liu2018partial,yu2019free,ma2019coarse,yu2020region}. Among them, Liu~\etal~\cite{Liu2018partial} attach heuristic mask update step to standard convolution and Yu~\etal~\cite{yu2019free} formally replace the mask update process with a learnable convolution layer. The second type is called valuable signals shifting that is inspired by the traditional exemplar-based approach~\cite{criminisi2003object}, which presently tends to model contextual attention to achieve~\cite{yan2018shift,yu2018generative,liu2019coherent,li2020recurrent,zeng2021cr}. In particular, RFR~\cite{li2020recurrent} applies multiple iterations at the bottleneck while sharing the attention scores to guide a patch-swap process.  ProFill~\cite{zeng2020high} iteratively performs inpainting based on the confidence map calculated by spatial attention. CRFill~\cite{zeng2021cr} yields a contextual reconstruction objective function that learns query-reference feature similarity. The third branch is to adopt auxiliary labels, which generate intermediate structures to assist with more accurate semantic~\cite{nazeri2019edgeconnect,liu2020rethinking,liao2021image,guo2021image}. In specific,  EC~\cite{nazeri2019edgeconnect} introduces canny edge to deliver finer inpainting structures. MEDFE~\cite{liu2020rethinking} jointly learns to represent structures and textures and utilizes spatial and channel equalization to ensure consistency. CTSDG~\cite{guo2021image} couples texture and structure through parallel pathways and then fuses them by bidirectional gated layers. In addition to the above methods, there also exist other approaches. One notable example is Score-SDE~\cite{DBLP:conf/iclr/0011SKKEP21} which proposes a scoring model that saves the gradient computation of energy-based models for efficient sampling.

\noindent
{\bf Inpainting with GAN Inversion.}
StyleGAN~\cite{karras2019style} implicitly learns hierarchical latent styles $w \in \mathbb{R}^{1\times512}$ instead of the initial stochastic vector $z$, which provides control over the style of outputs at coarse-to-fine levels of detail by style-modulation modules~\cite{huang2017arbitrary}. StyleGAN2~\cite{karras2020analyzing} further proposes weight demodulation, path length regularization, and generator redesign for improved image quality.
They are adept in the generation without any given images, but requires specialized networks~\cite{mao2017least} or regularization~\cite{gulrajani2017improved,miyato2018spectral} and paired training data. GAN inversion~\cite{zhu16generative} is a common practice that takes advantage of the intrinsic statistics of well-trained large-scale GAN as prior for generic applications~\cite{zhu2020domain,abdal2019image2stylegan}. Existing GAN inversion approaches could be roughly divided as optimized-based~\cite{cheng2021out,gu2020mganprior,wu2021stylespace,wang2021hijack} and encoder-based~\cite{richardson2021encoding,zhu2020domain,xu2021generative}. Among these methods, mGANprior~\cite{gu2020mganprior} utilizes multiple latent codes and adaptive channel importance for faithful reconstruction and shows applications in different tasks including inpainting. pSp~\cite{richardson2021encoding} synthesizes images with the mapping network to extract style vectors $w^+ \in \mathbb{R}^{18 \times 512}$ of latent space $\mathcal{W}^+$~\cite{abdal2019image2stylegan} separately for corresponding  $18$ style-modulation layers of the StyleGAN. Nevertheless, these approaches ignore the ``gapping'' issue, resulting in color inconsistency and semantic misalignment. 

\noindent
{\bf Difference with Previous Studies.}
In this paper we focus on encoder-based GAN inversion to improve generation fidelity for image inpainting. The proposed InvertFill is related to but significantly different from previous studies. In specific, InvertFill is relevant to the methods in~\cite{richardson2021encoding,zhu2020domain,xu2021generative} where encoder-based architecture is adopted. However, differing from them, we introduce a new $\mathcal{F}\&\mathcal{W}^+$ latent space to explicitly handle the ``gapping'' issue which is ignored in previous algorithms. Our method also shares similar spirit with the works of~\cite{zeng2020high,zeng2021cr} that adopt GAN for image inpainting. The difference is that these approaches may suffer from ambiguity when filling the large corruptions, while the proposed InvertFill exploits the priors of a large-scale generator and can achieve image inpainting with high-fidelity semantic.

\begin{figure*}[t]
	\centering
	\includegraphics[width=0.93\textwidth]{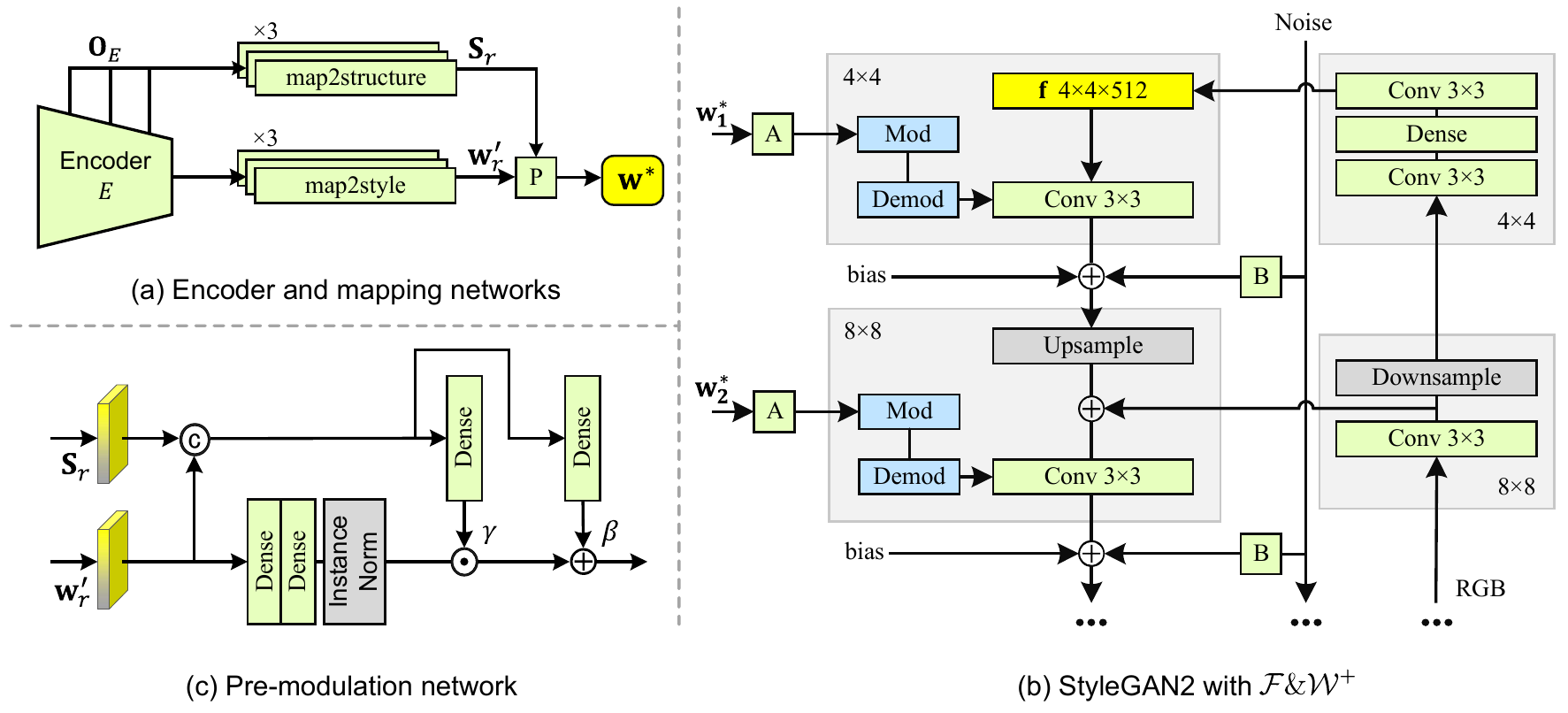}
	\caption{The main components of InvertFill, including a feature pyramid-based encoder (image (a)), mapping network with pre-modulation network (image (c)) and a StyleGAN2 generator with the proposed $\mathcal{F} \& \mathcal{W}^+$ latent space (image (b)). 
	}\label{fig:invertfill}
\end{figure*}

\section{The Proposed Method}

 Given an original image $\mathbf{I}$ and its corrupted image $\mathbf{I}_m = \mathbf{I} \odot (1 - \mathbf{M})$, where $\mathbf{M}$ is a binary mask and $\odot$ denotes element-wise product. The value of pixels in masked region $\mathbf{M}$ equal to 1 indicates invisible. We aim to produce a visually realistic reconstructed image $\mathbf{O}$ with the input of corrupted images $\mathbf{I}_m$. 
 

\subsection{$\mathcal{F} \& \mathcal{W}^+$ Latent Space}

Our architecture mainly consists of three components: (i) A feature pyramid-based~\cite{DBLP:conf/cvpr/LinDGHHB17} encoder $E$ that extracts input images and provides hierarchical reconstructed RGB images, (ii) the mapping networks with pre-modulation module, and (iii) a StyleGAN2 generator that takes in the style vectors as well as the input image $\mathbf{I}_m$ to generate a image. The details of InvertFill are shown in Fig.~\ref{fig:invertfill}. 

Specifically, we attach three RGB heads to the encoder $E$ for generating reconstructed RGB images $\mathbf{O}_{E}^r=\{\mathbf{O}_{E}^1,\mathbf{O}_{E}^2,\mathbf{O}_{E}^3\}$ in correspondence to three different scale. We follow the \textit{map2style}~\cite{richardson2021encoding} for the mapping network and cut down the network number from $18$ to $3$, each of which corresponds to the disentanglement level of image representation (i.e., coarse, middle and fine~\cite{karras2019style,richardson2021encoding}). Three \textit{map2style} networks encode the output feature map of the encoder into the intermediate code $\mathbf{w}^\prime \in \mathbb{R}^{3 \times 512}$. Similarly, we replicate \textit{map2style} as \textit{map2structure} to project reconstructed RGB images $\mathbf{O}_{E}^r$ gradually into structure vector $\mathbf{S}_r = \{\mathbf{S}_{1}, \mathbf{S}_{2}, \mathbf{S}_{3}\}$.

Before executing the style modulation in the generator, we perform $L$ pre-modulation networks to project the semantic structure $\mathbf{S}_r$ into the style vector $\mathbf{w}^*$ in latent space $\mathcal{F} \& \mathcal{W}^+$, i.e., $\mathbf{w}^* = E(\mathbf{I}_m), \mathbf{w}^* \in \mathbb{R}^{L \times 512}$. $L = \log_2(s) \cdot 2 - 2$ denotes number of style-modulation layers of StyleGAN2 generator, and is adjusted by the image resolution $s$ on the generator side. As Fig.~\ref{fig:invertfill}(c) demonstrates, we adopt Instance Normalization (IN)~\cite{ulyanov2016instance} to regularize the $\mathbf{w}^\prime$ latent code, then carry out denormalization according to multi-scale structure vector $\mathbf{S}_r$, 
\begin{equation}
    \mathbf{w}_{l}^* = \gamma \odot \text{IN}\left(\mathbf{w}^{\prime}_{r}\right) + \beta ,
\end{equation}
where $l \in \{1,2, \dots ,L\}$ denotes the index of style vectors, $r \in [1,3]$ indicates three vectors $\mathbf{w}^{\prime}$ correspond to level of coarse to fine, $(\gamma, \beta)$ is a pair of the affine transformation parameters learned by networks shown in Fig.~\ref{fig:invertfill}(c). Different than previous methods in only using intermediate latent code from a network, the proposed pre-modulation module is a lightweight network and novel in applying more discriminative multi-scale features to help latent code perceive uncorrupted prior and better guide image generation.


The GAN is initially fed with a stochastic vector $z \in \mathcal{Z}$, and previous works~\cite{abdal2019image2stylegan,gu2020mganprior,richardson2021encoding,zhu2020domain} invert the source images into the intermediate latent space $\mathcal{W}$ or $\mathcal{W}^+$,  which is a less entangled representation than latent space $\mathcal{Z}$. The style vectors $w \in \mathcal{W}$ or $w^+ \in \mathcal{W}^+$ are sent to the style-modulation layers of pre-trained StyleGAN2 to synthesize target images. These approaches can be formulated mathematically as follows,
\begin{equation}
	\mathbf{O}_G = G(E(\mathbf{I}_m)),E(\mathbf{I}_m) \sim W^+ ,
	\label{equ:ori}
\end{equation}
where $E(\cdot)$ and $G(\cdot)$ represent the encoder that maps source images into latent space and the pre-trained GAN generator, respectively. 

Nevertheless, the above formulation in Equ.~(\ref{equ:ori}) may encounter the ``gapping'' issue in image translation tasks with hard constraint, \eg, image inpainting. The hard constraint requires that parts of the source and recovered image remain the same. We formally defined the hard constraint in image inpainting as $\mathbf{I} \odot (1-\mathbf{M}) \equiv \mathbf{O} \odot (1 - \mathbf{M})$. Intuitively, we argue that the ``gapping'' issue is caused by that the GAN model cannot directly access pixels of the input image but the intermediate latent code. To avoid the semantic inconsistency and color discrepancy caused by this problem, we utilize the corrupted image $\mathbf{I}_m$ as one of the inputs to assist with the GAN generator inspired by skip connection of U-Net~\cite{ronneberger2015u}. In detail, $\mathbf{I}_m$ is fed into the RGB branch as shown in Fig.~\ref{fig:invertfill}(b), the feature map between RGB branch and the generator are connected by element-wise addition.
Hence, the previous formulation in Equ. \eqref{equ:ori} is updated as:
\begin{equation}
    \mathbf{O}_G = G(E(\mathbf{I}_m), \mathbf{I}_m), E(\mathbf{I}_m) \sim \mathcal{F} \& \mathcal{W}^+.
\end{equation}


\subsection{Soft-update Mean Latent}

Pixels closer to the mask boundary are more accessible to inpainting, but conversely the model is hard to predict specific content missing. We find that the encoder learns a trick to averaging textures to reconstruct the region away from unmasked region. It causes blurring or mosaic in some areas of the output image, mainly located away from the mask borders, as shown in Fig~\ref{fig:RQ4-SML}. Drawing inspiration from L2 regularization and motivated by the intuition that fitting diverse domains works better than fitting a preset static domain, a feasible solution is to make style code $\mathbf{w}^*$ be bounded by the mean latent code of pre-trained GAN.

The mean latent code is obtained from abundant random samples that restrict the encoder outputs to the average style hence lossy the diversity of output distribution of encoder. In addition, it introduces additional hyperparameters and a static mean latent code that requires loading when training the model.

We adopt dynamic mean latent code instead of static one by stochastically fluctuating the mean latent code while training. 
Further, we smooth the effect of fluctuating variance for convergence inspired by a reinforcement learning~\cite{DBLP:journals/corr/LillicrapHPHETS15}. 
For initialization, target mean latent code $\mathbf{\overline{w}}_t$ and online mean latent code $\mathbf{\overline{w}}_o$ are sampled. 
$\overline{\mathbf{w}}_o$ is used in image generation instead of $\overline{\mathbf{w}}_{t}$, which is fixed until $\overline{\mathbf{w}}_o = \overline{\mathbf{w}}_t$ and then resampled. 
Between two successive sampled mean latent codes, $\overline{\mathbf{w}}_o$ is updated by $\mathbf{\overline{w}}_o \leftarrow \tau \mathbf{\overline{w}}_o + (1- \tau) \mathbf{\overline{w}}_t$ per iteration during training, where $\tau$ denotes updating factor and $\mathbf{\overline{w}}_t $ for soft updating target mean latent code. The soft-update mean latent degraded to static mean latent~\cite{richardson2021encoding} when the parameter $\tau$ of soft-update mean latent approaching zero. 

\subsection{Optimization}

\begin{figure*}[t]
	\begin{overpic}[width=\textwidth]{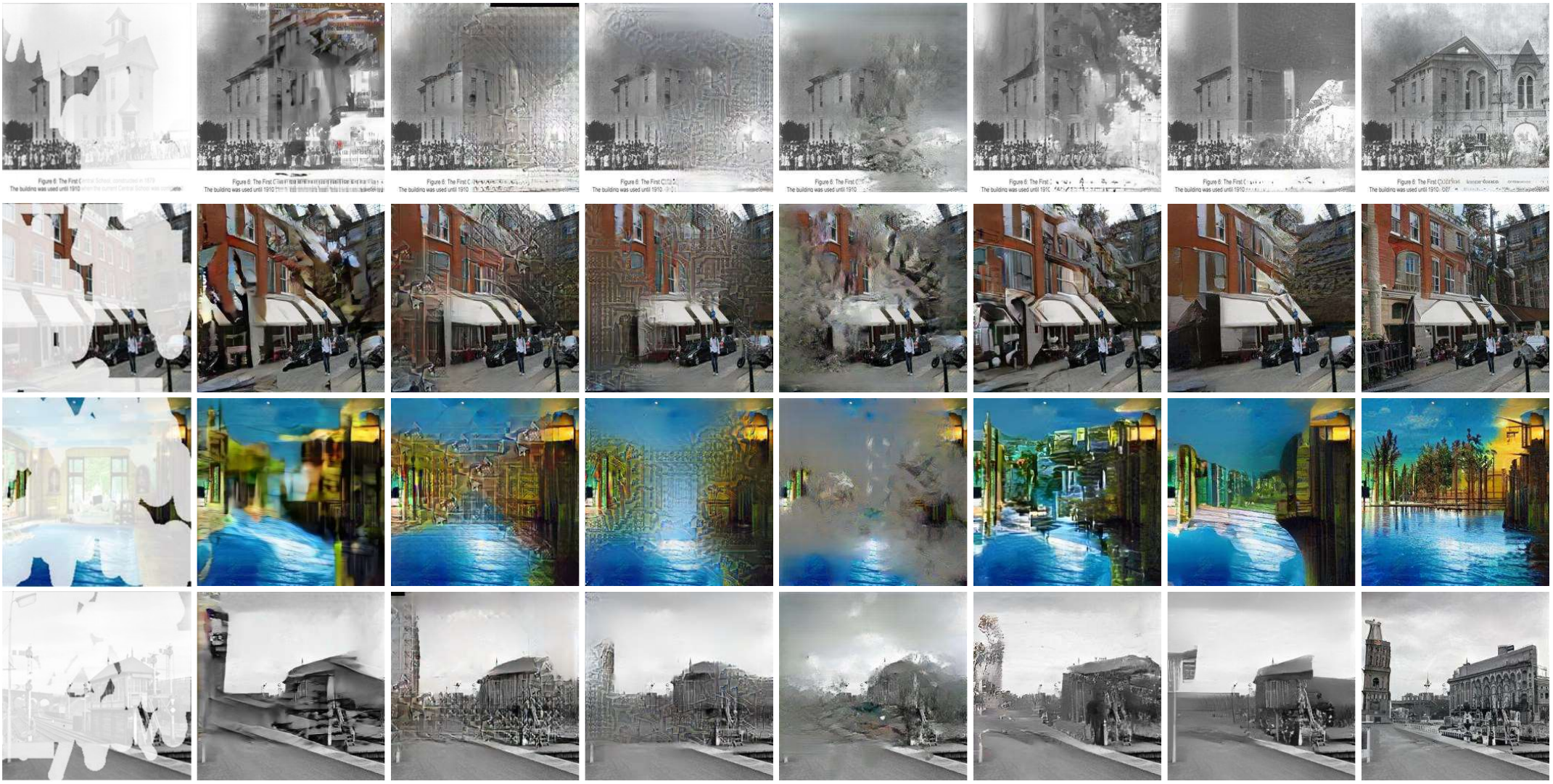} \small
		\put(2,-2){Masked}
		\put(16.5,-2){GC}
		\put(28.5,-2){RFR}
		\put(39,-2){CTSDG}
		\put(51,-2){MEDFE}
		\put(64.5,-2){CRFill}
		\put(77,-2){ProFill}
		\put(90.5,-2){Ours}
	\end{overpic}
	\caption{Qualitative results on Places2 dataset.}\label{fig:QualiComparison_Places2}
\end{figure*}

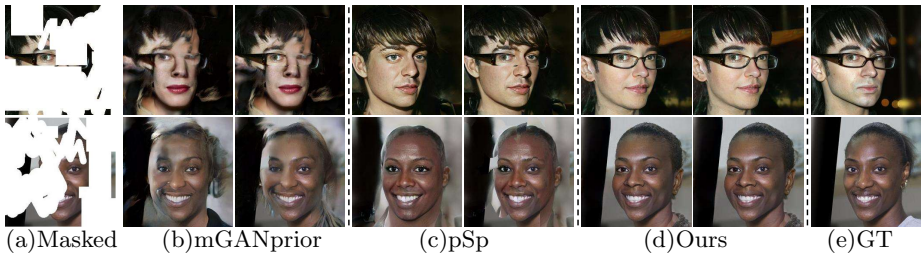
\begin{figure*}[t]
    \tiny
    \centering
	\begin{overpic}[width=\textwidth]{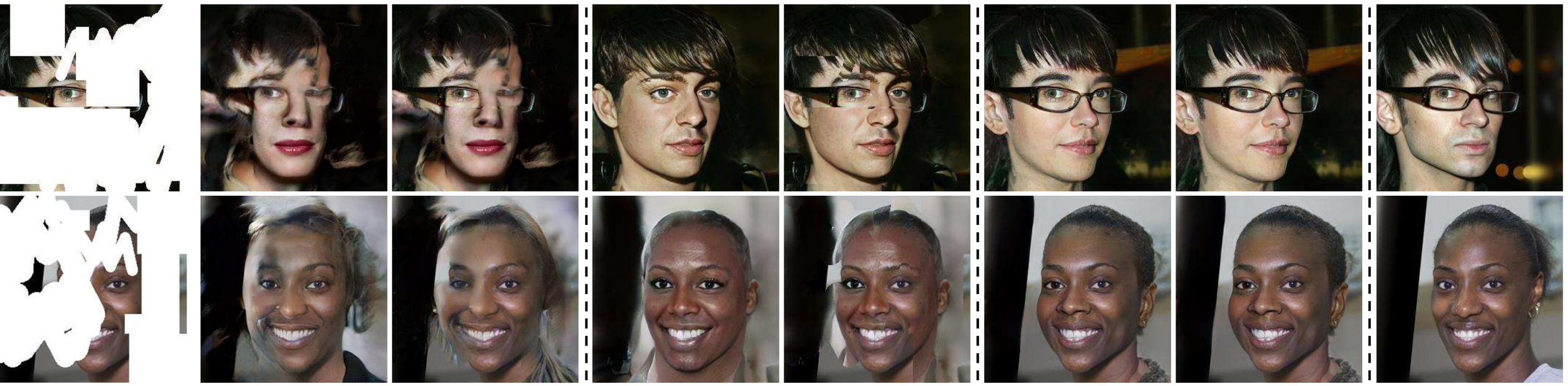} \small
		\put(0,-2){(a)Masked}
		\put(17,-2){(b)mGANprior}
		\put(45,-2){(c)pSp}
		\put(69.5,-2){(d)Ours}
		\put(89.5,-2){(e)GT}
	\end{overpic}
	\caption{Qualitative results on CelebA-HQ dataset. Two columns of (b-d) show the original model output and composition output, from left to right, respectively. The output of the GAN-inversion-based method (pSp~\cite{richardson2021encoding} and mGANprior~\cite{gu2020mganprior}) is inconsistency at the edge of the mask. Zoom-in to see the details.}\label{fig:QualiComparison_CHQ}
\end{figure*}

Following prior work in inpainting~\cite{Liu2018partial,li2020recurrent}, our architecture is supervised by regular inpainting loss $\mathcal{L}_{\text{ipt}}$, which consists of the pixel-wise Euclidean norm of valid and hole regions, the perceptual loss $\mathcal{\text{perc}}$, the style loss $\mathcal{\text{style}}$, and the total variation loss $\mathcal{\text{tv}}$:
\begin{equation}
    \mathcal{L}_{\text{ipt}} = \mathcal{L}_{\text{valid}} + \mathcal{L}_{\text{hole}} +  \mathcal{L}_\text{perc} + \mathcal{L}_\text{style} + \mathcal{L}_\text{tv} ,
    \label{eq:ipt}
\end{equation}
where all above distance are calculated between $\mathbf{I}$ and $\mathbf{O}_G$.  $\mathcal{L}_{\text{valid}}$ and $\mathcal{L}_{\text{hole}}$ are $\ell_1$ norm on the known and masked region respectively. 
The perceptual loss $\mathcal{L}_{\text{perc}}$ and the style loss $\mathcal{L}_{\text{style}}$ are based on a pre-trained VGG-16 network. More details can be found in~\cite{li2020recurrent}.

To directly optimize our encoder, the multi-scale reconstruction loss $\mathcal{L}_{\text{msr}}$ is utilized to penalize the deviation of $\mathbf{O}_{E}^{r}$ at each scale:
\begin{equation}
    \mathcal{L}_{\text{msr}} = \sum_{r=1}^{3} (\mathcal{L}_{\text{perc}}^{r}
    + \mathcal{L}_{\text{style}}^{r} + \mathcal{L}_{\text{rec}}^r) ,
    \label{eq:msr}
\end{equation}
where $\mathcal{L}_{\text{rec}}$ is represented as mean-squared loss between $\mathbf{I}$ and $\mathbf{O}_E$. The multi-scale reconstruction loss $\mathcal{L}_{\mathrm{msr}}$ contains three different losses including perceptual ($\mathcal{L}_{\mathrm{perc}}^{r}$)~\cite{DBLP:journals/corr/GatysEB15a}, style ($\mathcal{L}_{\mathrm{style}}^{r}$)~\cite{Liu2018partial} and mean-square ($\mathcal{L}_{\mathrm{rec}}^{r}$) losses. The role of $\mathcal{L}_{\mathrm{msr}}$ is to supervise the generated image from decoder and make final generation close to the original image.

The soft-update mean latent is utilized to prevent the encoder from falling into the trick way. We adopt the following fidelity loss $\mathcal{L}_{\text{fid}}$ for improving the quality and diversity of output images:
\begin{equation}
    \mathcal{L}_{\text{fid}} = \| \mathbf{w}^* - \overline{\mathbf{w}}_o \|_2, \text{ } (\mathbf{w}^*, \overline{\mathbf{w}}_o) \in \mathbb{R}^{L \times 512} .
\end{equation}

The fidelity loss $\mathcal{L}_{\mathrm{fid}}$ is designed as a mean squared loss of style vectors $\mathrm{\mathbf{w}}^{*}$ and online mean latent code $\mathrm{\bar{\mathbf{w}}}_o$. Its role is to improve the quality and diversity of the output images.

Overall, the loss of our networks is defined as the weighted sum of the inpainting loss, the multi-scale reconstruction loss, and the fidelity loss.
\begin{equation}
    \mathcal{L} = \mathcal{L}_{\text{ipt}} + \lambda_{\text{msr}} \mathcal{L}_{\text{msr}} + \lambda_{\text{fid}} \mathcal{L}_{\text{fid}} ,
    \label{eq:all}
\end{equation}
where $\lambda_{\text{msr}}$ and $\lambda_{\text{fid}}$ are the balancing factors for the multi-scale reconstruction loss and the fidelity loss, respectively.

\section{Experiments}

\begin{figure*}[t]
   \begin{overpic}[width=\textwidth]{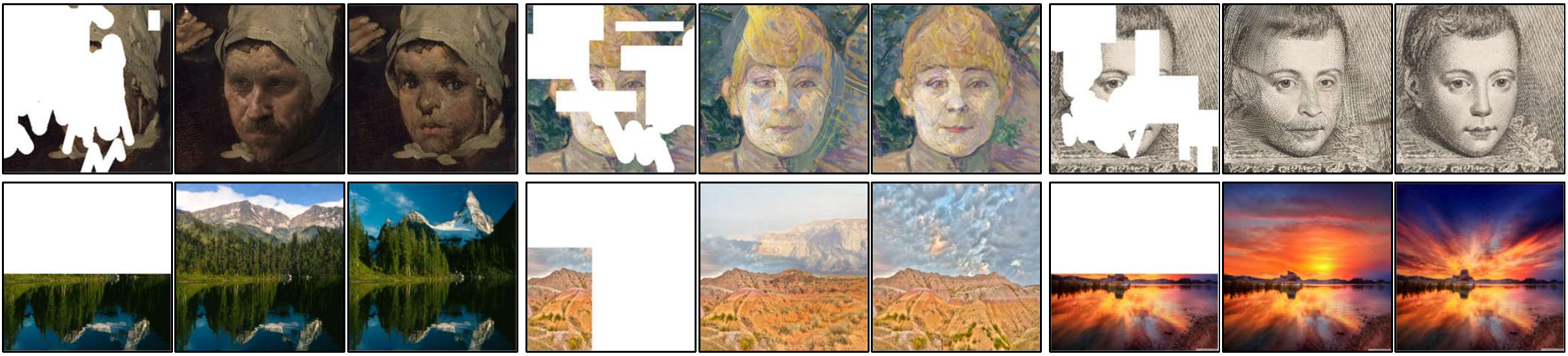} \small
   \end{overpic}
   \caption{The visual effect of our method for processing input images from unseen domain. The 1st row shows inpainting results of Metfaces, and the 2nd row shows outpainting results of Scenery. Each instance of results is laid out as the masked image, the model output, and the original image.}\label{fig:RQ3}
\end{figure*}

We perform extensive validating experiments aiming to answer the following research questions: 
\begin{itemize}
    \item \textbf{RQ1:} How does our approach perform,  compared with existing methods, especially the fidelity when the input is large-scale masked images.
    \item \textbf{RQ2:} Can our approach resolve the ``gapping'' issue?
    \item \textbf{RQ3:} Can our approach handle input from unseen domain by reusing the well-trained generator while only retraining a lightweight encoder?
    \item \textbf{RQ4:} How do different components (e.g., soft-update mean latent, pre-modulation) affect our approach? 
\end{itemize}

\subsection{Experimental Settings}
{\flushleft {\bf Datasets.}} Experiments for RQ1, RQ2 and RQ4 are conducted on two datasets, Places2~\cite{zhou2017places} and CelebA-HQ~\cite{karras2018progressive}. CelebA-HQ contains 30,000 high-resolution celebrity faces, and we follow~\cite{yu2018generative,yu2019free} to split this dataset for training and testing. Places2 contains real-world photos, including more significant objects, such as streets, cars, houses, which is better suited for verifying models on large-scale masks than CelebA-HQ. Based on the official \textit{train}/\textit{val}/\textit{test} split, we train the model on \textit{train} plus \textit{test} about 200,000 images, evaluate the model on first 5,000 images of \textit{val}. With regard to RQ3, we utilize two datasets Scenery~\cite{yang2019very} and MetFaces~\cite{DBLP:conf/nips/KarrasAHLLA20}. Scenery dataset is a common benchmark for recent image outpainting tasks and contains 6,040 landscape photographs. We follow~\cite{yang2019very} to use about 5,000 images as training set and the remaining 1,000 images as test set. MetFaces consists of 1,336 human faces extracted from works of art, and we randomly select 1,000 images as training set and other images as test set. Our model and all baselines adopt the same training and test strategies to ensure experimental fairness.

\begin{table}[t]
	\small
    \centering
\caption{Quantitative comparison with the mainstream inpainting approaches on Places2 and CelebA-HQ datasets. \textit{Hard}, \textit{Extreme}, \textit{All} masks denote the mask with coverage ratio of 50\% $\sim$ 60\%, 70\% $\sim$ 90\%, and 10\% $\sim$ 90\% , respectively. $\uparrow$ Higher is better, and $\downarrow$ lower is better. Best and second best results are \textbf{highlighted}.}
\label{tab:QuanComparison}
\setlength{\tabcolsep}{3.5pt}
\begin{tabular}{lclccccccc}
\hline
 &
  \multicolumn{1}{l}{} &
   &
  GC &
  RFR &
  MEDFE &
  ProFill &
  CTSDG &
  CRFill &
  Ours \\ \hline
\multicolumn{1}{l|}{\multirow{9}{*}{\textbf{\rotatebox{90}{Places2}}}} &
  \multirow{3}{*}{\textit{\textbf{hard}}} &
  SSIM$\uparrow$ &
  0.624 &
  0.645 &
  0.598 &
  \textbf{0.664} &
  0.651 &
  0.629 &
  0.641 \\
\multicolumn{1}{l|}{} &
   &
  FID$\downarrow$ &
  22.05 &
  27.77 &
  44.38 &
  21.49 &
  35.77 &
  22.46 &
  \textbf{12.44} \\
\multicolumn{1}{l|}{} &
   &
  LPIPS$\downarrow$ &
  0.246 &
  0.235 &
  0.294 &
  0.240 &
  0.272 &
  0.250 &
  \textbf{0.232} \\ \cline{2-10} 
\multicolumn{1}{l|}{} &
  \multirow{3}{*}{\textit{\textbf{extreme}}} &
  SSIM$\uparrow$ &
  0.363 &
  0.382 &
  0.323 &
  \textbf{0.409} &
  0.393 &
  0.360 &
  0.366 \\
\multicolumn{1}{l|}{} &
   &
  FID$\downarrow$ &
  51.35 &
  71.19 &
  111.85 &
  46.44 &
  95.50 &
  51.26 &
  \textbf{21.08} \\
\multicolumn{1}{l|}{} &
   &
  LPIPS$\downarrow$ &
  0.407 &
  0.395 &
  0.495 &
  0.402 &
  0.438 &
  0.413 &
  \textbf{0.386} \\ \cline{2-10} 
\multicolumn{1}{l|}{} &
  \multirow{3}{*}{\textit{\textbf{all}}} &
  SSIM$\uparrow$ &
  0.734 &
  0.750 &
  0.714 &
  \textbf{0.764} &
  0.755 &
  0.738 &
  0.761 \\
\multicolumn{1}{l|}{} &
   &
  FID$\downarrow$ &
  14.19 &
  16.26 &
  26.15 &
  13.81 &
  21.36 &
  14.44 &
  \textbf{9.29} \\
\multicolumn{1}{l|}{} &
   &
  LPIPS$\downarrow$ &
  0.178 &
  0.170 &
  0.217 &
  0.173 &
  0.199 &
  0.182 &
  \textbf{0.155} \\ \hline\hline
\multicolumn{1}{l|}{\multirow{9}{*}{\textbf{\rotatebox{90}{CelebA-HQ}}}} &
  \multirow{3}{*}{\textit{\textbf{hard}}} &
  SSIM$\uparrow$ &
  0.790 &
  \textbf{0.825} &
  0.781 &
  - &
  0.818 &
  0.810 &
  0.812 \\
\multicolumn{1}{l|}{} &
   &
  FID$\downarrow$ &
  17.38 &
  9.98 &
  21.97 &
  - &
  15.13 &
  13.78 &
  \textbf{9.89} \\
\multicolumn{1}{l|}{} &
   &
  LPIPS$\downarrow$ &
  0.170 &
  0.128 &
  0.192 &
  - &
  0.151 &
  0.139 &
  \textbf{0.121} \\ \cline{2-10} 
\multicolumn{1}{l|}{} &
  \multirow{3}{*}{\textit{\textbf{extreme}}} &
  SSIM$\uparrow$ &
  0.589 &
  0.641 &
  0.552 &
  - &
  0.616 &
  0.639 &
  \textbf{0.652} \\
\multicolumn{1}{l|}{} &
   &
  FID$\downarrow$ &
  41.70 &
  22.07 &
  55.52 &
  - &
  33.89 &
  30.19 &
  \textbf{13.21} \\
\multicolumn{1}{l|}{} &
   &
  LPIPS$\downarrow$ &
  0.297 &
  0.241 &
  0.359 &
  - &
  0.281 &
  0.275 &
  \textbf{0.214} \\ \cline{2-10} 
\multicolumn{1}{l|}{} &
  \multirow{3}{*}{\textit{\textbf{all}}} &
  SSIM$\uparrow$ &
  0.852 &
  \textbf{0.878} &
  0.846 &
  - &
  0.875 &
  0.859 &
  0.867 \\
\multicolumn{1}{l|}{} &
   &
  FID$\downarrow$ &
  11.78 &
  7.96 &
  15.52 &
  - &
  10.32 &
  11.94 &
  \textbf{7.71} \\
\multicolumn{1}{l|}{} &
   &
  LPIPS$\downarrow$ &
  0.128 &
  0.092 &
  0.142 &
  - &
  0.110 &
  0.114 &
  \textbf{0.089} \\ \hline
\end{tabular}
\end{table}

{\flushleft {\bf Evaluation Metrics.}} We use three metrics following prior works to measure the quality and fidelity of inpainting results. SSIM~\cite{wang2004image} modeling image distortion by structure, luminance, and contrast, is a pixel-level objective metric similar to PSNR, and their drawbacks cause inconsistent evaluation results with the human eye. Despite that, they are classical metrics for image evaluation, one of which SSIM we selected for quantitative comparison. FID~\cite{heusel2017gans} is a deep metric and closer to human perception. It measures the distribution distance with a pre-trained inception model, which better captures distortions. LPIPS~\cite{zhang2018unreasonable} is another learned perceptual metric and commonly used to score the intra-conditioning diversity of models output. Following previous works~\cite{li2020recurrent,nazeri2019edgeconnect,yu2019free}, We calculate these quantitative metrics on original images $\mathbf{I}$ and composition images $\mathbf{I} \odot (1 - \mathbf{M}) + \mathbf{O}_{G} \odot \mathbf{M}$.

{\flushleft {\bf Baselines.}} We carefully select baseline methods mainly from two perspectives: UNet style methods and Inversion style methods to demonstrate our approach's characteristics and superiority. First, for the sake of validating the ability of InvertFill in filling images under large-scale masks, we compare it with the previous approaches including EC~\cite{nazeri2019edgeconnect}, GC~\cite{yu2018generative}, RFR~\cite{li2020recurrent}, MEDFE~\cite{liu2020rethinking}, ProFill~\cite{zeng2020high}, CTSDG~\cite{guo2021image} and CRFill~\cite{zeng2021cr}. Second, we compare with the latest GAN inversion-based inpainting methods mGANprior~\cite{gu2020mganprior} and pSp~\cite{richardson2021encoding}.

\subsection{Implementation Details}

We utilize eight A100 GPUs for pre-training the GAN generator, and one TITAN RTX GPU for optimizing the encoder and other experiments. Following~\cite{li2020recurrent}, we scale the image size of all datasets to $256 \times 256$ as the input. In the light of the mask coverage, we classify the test masks into three difficulty levels: \textit{Hard}/\textit{Extreme}/\textit{All}, indicates the mask with coverage ratio of 50\% $\sim$ 60\%, 70\% $\sim$ 90\%, 10\% $\sim$ 90\%, respectively. During testing, for a fair comparison, we use the same image-mask pair for all approaches. More details of implementation are shown in the supplementary.

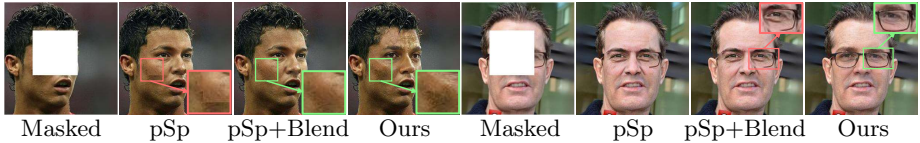
\begin{figure}[t]
    \centering
   \begin{overpic}[width=\columnwidth]{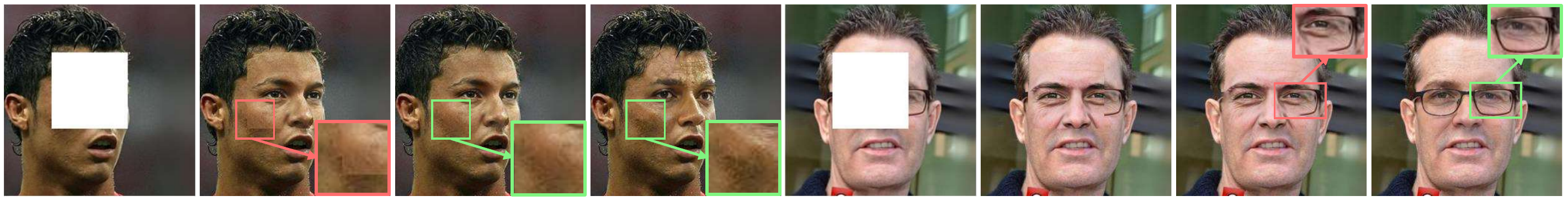} \small
	\put(2,-2){Masked}
	\put(16,-2){pSp}
	\put(24.5,-2){pSp+Blend}
	\put(41,-2){Ours}
	\put(52,-2){Masked}
	\put(66.5,-2){pSp}
	\put(74.3,-2){pSp+Blend}
	\put(91,-2){Ours}
	\end{overpic}
  \caption{
  Comparison with pSp~\cite{richardson2021encoding} and pSp+Blend~\cite{wu2021gpgan} that post-processing by image blending. The 1st row shows the color discrepancy that image blending is sufficient to resolve satisfactorily. The 2nd row shows that the semantic inconsistency is still reserved, except for our method.}\label{fig:RQ2}
\end{figure}

\subsection{Result Analysis}

{\flushleft {\bf RQ1.}} We reproduce all the above baselines by utilizing their official implementations. Concerning Places2 dataset, we utilize the pre-trained weights officially released by the baselines. On CelebA-HQ dataset, EC~\cite{nazeri2019edgeconnect}, GC~\cite{yu2019free}, mGANprior~\cite{gu2020mganprior} offer pre-trained weights, we thereby carefully retrain other baselines through the official source codes. Because ProFill only offers Web API on Places2, we use placeholder `-' in Table~\ref{tab:QuanComparison} for ProFill on CelebA-HQ.

From Table~\ref{tab:QuanComparison}, our method achieves the best or comparable performance among advanced inpainting approaches. In terms of the FID metric, our method at most produces a notable margin of 54.60\% and 40.14\% on Places2 and CelebA-HQ datasets, respectively. And our method also outperforms the second-best approach 11.2\% and 10.4\% improvements on another perceptual metric LPIPS, which validates the superiority of our design.


\begin{table}[t]
\begin{minipage}[t]{0.48\textwidth}
    \makeatletter\def\@captype{table}
    \scriptsize
    \caption{Comparison with previous GAN inversion-based and diffusion-based approaches on CelebA-HQ dataset.}
    \label{tab:RQ2}
\resizebox{\columnwidth}{12mm}{
\begin{tabular}{lccc}
\hline
            & FID$\downarrow$ & LPIPS$\downarrow$ & SSIM$\uparrow$  \\ \hline\hline
Score-SDE~\cite{DBLP:conf/iclr/0011SKKEP21}   &  24.76 & 0.337  & 0.428 \\
mGANprior~\cite{gu2020mganprior} &  29.57 & 0.273  & 0.608 \\
pSp~\cite{richardson2021encoding}   & 25.61 & 0.248 & 0.594  \\
pSp + Blend~\cite{wu2021gpgan} & 21.96 & 0.240 & 0.602    \\
Ours        & \textbf{13.21} & \textbf{0.214} & \textbf{0.652}  \\ \hline
\end{tabular}
}
    \end{minipage}
\quad 
\begin{minipage}[t]{0.48\textwidth}
    \makeatletter\def\@captype{table}
	\caption{Comparison with previous outpainting approaches and inpainting baselines on Scenery dataset.}
	\label{tab:RQ3}
    \tiny
    \centering
    \resizebox{\columnwidth}{12mm}{
	\begin{tabular}{llll}
		\hline
		& FID$\downarrow$ & LPIPS$\downarrow$ & SSIM$\uparrow$ \\ \hline\hline
		RFR~\cite{li2020recurrent}         & 138.31  &  0.455 &  0.376  \\
		pSp~\cite{richardson2021encoding}  &  49.62  & 0.379 &    0.392   \\ 
		Boundless~\cite{krishnan2019boundless}     & 45.05  &  0.368 &  0.413 \\
		NS-outpaint~\cite{yang2019very}  &  38.95  &  0.342  &  0.410   \\ 
		Ours        &  \textbf{20.90}   & \textbf{0.294}  &   \textbf{0.439}  \\ \hline
	\end{tabular}}
    \end{minipage}
\end{table}

Fig.~\ref{fig:QualiComparison_Places2} and~\ref{fig:QualiComparison_CHQ} provide several visual inpainting results on Places2 and CelebA-HQ datasets. Fig.~\ref{fig:QualiComparison_Places2} reveals that the prior works still struggle to generate refined texture if the input image with large corruptions, while our approach has been able to create semantically rich objects such as windows, towers, and woods. In Fig.~\ref{fig:QualiComparison_CHQ}, mGANprior~\cite{gu2020mganprior} progressively erases the color discrepancy rely on optimized-based inversion but is unable to bypass semantic inconsistency. The encoder-based inversion method pSp~\cite{richardson2021encoding} could synthesize realistic pixels for corrupted regions based on the well-trained model, though it still has not resolved the ``gapping'' issue. The results indicate that our method produces consistent output while generating high-fidelity texture compared to existing methods.

{\flushleft {\bf RQ2.}} The ``gapping'' causes color discrepancy and semantic inconsistency, and we are counting on image post-processing to tackle this issue at the beginning of this study. Specifically, we adopt image blending~\cite{wu2021gpgan}, which is effective in eliminating the color discrepancy but helpful in remedying semantic inconsistency.

To further demonstrate the superiority of our method, we construct the pSp+Blend variant that introduces an image blending~\cite{wu2021gpgan} method after generating output images. In Fig.~\ref{fig:RQ2}, the first row shows the distinct gap at the stitching boundary in pSp output, and pSp+Blend fixes this color discrepancy problem. Even so, the second row shows pSp+Blend unable to assist with semantic inconsistency problem given the glasses are still incomplete. Compared with the vanilla pSp and pSp+Blend, output images of our approach no longer suffer from color discrepancy or semantic inconsistency. 

We conduct a comparison experiment on CelebA-HQ dataset with the \textit{Extreme} level masks. As demonstrated in Table~\ref{tab:RQ2}, our method performs better than a recent diffusion-based approach Score-SDE~\cite{DBLP:conf/iclr/0011SKKEP21} w.r.t to FID, LPIPS and SSIM metrics. The results in Table~\ref{tab:RQ2} also show that our method performs best among the existing inversion-based inpainting approaches after resolving the ``gapping'' issue. Notably, our method does not require any image post-processing. 

\begin{table}[t]
\centering
\scriptsize
\caption{Comparison with previous inpainting methods on Metfaces. In this experimental setting, the model/generator is only trained on CelebA-HQ.}
\label{tab:RQ3-Metfaces}
\setlength{\tabcolsep}{2.4mm}{
\begin{tabular}{c|lll|lll}
\hline
\multirow{2}{*}{Method} & \multicolumn{3}{c|}{Easy} & \multicolumn{3}{c}{Extreme} \\ \cline{2-7} 
                        & SSIM$\uparrow$   & FID$\downarrow$    & LPIPS$\downarrow$   & SSIM$\uparrow$    & FID$\downarrow$     & LPIPS$\downarrow$   \\ \hline\hline
RFR & 0.93   & 18.89   & 0.069   & 0.52    & 58.24  & 0.315   \\
CRFill & 0.95   & 13.67   & 0.042  & 0.54  & 50.93   &  0.278       \\\hline
pSp & 0.95  & 14.91   & 0.040  & 0.49  & 65.04   & 0.341   \\
Ours & \textbf{0.97}   & \textbf{8.64}  & \textbf{0.033}  & \textbf{0.60}   & \textbf{38.85}   & \textbf{0.227}   \\ \hline
\end{tabular}}
\end{table}

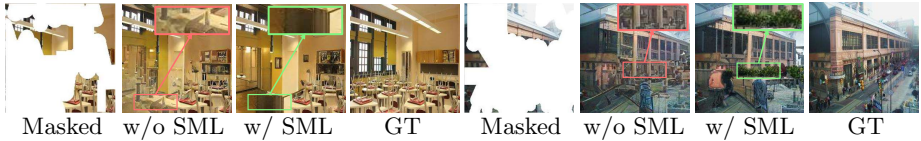
\begin{figure}[t]
	\small
	\centering
   \begin{overpic}[width=\columnwidth]{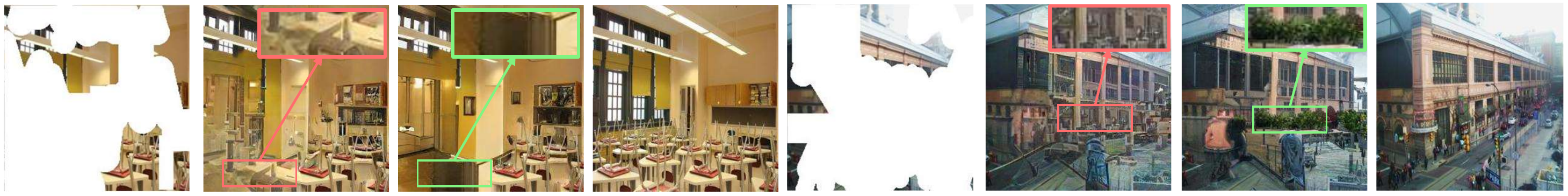} \small
	\put(2,-2){Masked}
	\put(13,-2){w/o SML}
	\put(26.2,-2){w/ SML}
	\put(41.5,-2){GT}
	\put(52,-2){Masked}
	\put(63.2,-2){w/o SML}
	\put(76.5,-2){w/ SML}
	\put(92,-2){GT}
	\end{overpic}
  \caption{
   The importance of soft-update mean latent.
  }\label{fig:RQ4-SML}
\end{figure}

{\flushleft {\bf RQ3.}} Concerning validating that our approach can reuse the pre-trained GAN generator as priors to tackle image from unseen domain, we conduct two extended tests that introduced images or masks from unseen domains and only required optimizing the lightweight encoder. The first is archaic photograph inpainting, and we use MetFaces~\cite{DBLP:conf/nips/KarrasAHLLA20} for optimizing the encoder, and remain the pre-trained weights of GAN generator of CelebA-HQ dataset. For the second one, we perform our approach with outpainting masks~\cite{krishnan2019boundless} on Scenery dataset. Similarly, the generator did not retrain on the Scenery dataset rather than remaining the weights for Places2.

The 1st row of Fig.~\ref{fig:RQ3} shows the inpainting results of archaic photograph inpainting. It demonstrates that our method enables the generator to synthesize semantically consistent style and patches, even in an unseen domain. From the 2nd row of Fig.~\ref{fig:RQ3}, the outpainting results on the Scenery dataset show our approach still can synthesize realistic texture and significant objects, \eg trees, mountains. To ensure the masks are unseen for the GAN generator, we only use the outpainting masks to train the encoder, not the GAN generator. 

Furthermore, we quantitatively compare mainstream outpainting approaches as well as adopt RFR~\cite{li2020recurrent} and pSp~\cite{richardson2021encoding} as additional baselines. As shown in Table~\ref{tab:RQ3}, our model considerably outperforms the best outpainting baselines~\cite{krishnan2019boundless,yang2019very} with respect to FID, LPIPS, and SSIM. Similarly, we conduct experiments compared with inpainting baselines on Metfaces, as show in Table~\ref{tab:RQ3-Metfaces}. In summary, the results indicate that our proposed method is robust and extends to other tasks with out-of-domain inputs.

Due to limited space, please kindly refer to the supplementary material for more results.

\subsection{Ablation Study (RQ4)}

The ablation experiments are carried out on the Places2 dataset under the \textit{Extreme} mask setting. 
In Table~\ref{tab:RQ4}, we construct three variants to verify the contribution of proposed modules, in which PM and SML denote pre-modulation and soft-update mean latent. 
By learning from these modules, our method considerably outperforms the most naive variant w.r.t FID, LPIPS, SSIM, and PSNR.

The soft-update mean latent is motivated by the intuition that fitting diverse domains works better than fitting a preset static domain, especially when the training dataset contains various scenarios such as street and landscape. As shown in Fig.~\ref{fig:RQ4-SML}, when we use SML code that dynamically fluctuates during training, the masked region far away from the mask border tends to be reconstructed by explicitly learned semantics instead of repetitive patterns. Notably, `w/o SML' represents using regular static mean latent code.

\subsection{Failure Cases and Discussion}

Fig.~\ref{fig:fail} shows two failure cases.
Even if the model can recognize the corrupted objects (our method tends to recover the human face in the left case of Fig.~\ref{fig:fail}), it mistakenly locates them and produces severe artifacts.
When lacking sufficient prior knowledge, our method fails to reconstruct details. 
This demonstrates that these situations are challenging for image inpainting and need further study.

\begin{table}[t]
\centering
\caption{Ablation study comparison on Places2 dataset under \textit{Extreme} mask setting.}
\label{tab:RQ4}
\setlength{\tabcolsep}{8pt}
\begin{tabular}{cccc|cccc}
\hline
$\mathcal{F} \& \mathcal{W}^+$ & SML & PM & $\mathcal{W}^+$ & FID & LPIPS & SSIM & PSNR \\
\hline\hline
\checkmark  &       &     &    &   35.37   & 0.395 & 0.357 & 13.85     \\
\checkmark   & \checkmark     &     &    &  24.73   &    0.389  & 0.358 & 13.99    \\
\checkmark   & \checkmark     & \checkmark   &    & 21.08 & 0.386  & 0.366   & 14.62  \\
   & \checkmark     & \checkmark   & \checkmark  &  42.85   &  0.392   &   0.361   &  14.25  \\\hline
\end{tabular}
\end{table}

\newcommand{\M}{0.155}
\renewcommand\arraystretch{0.8}
\begin{figure}[t]
	\centering
	\begin{tabular}{cccccc}
	\frame{\includegraphics[width=\M\linewidth]{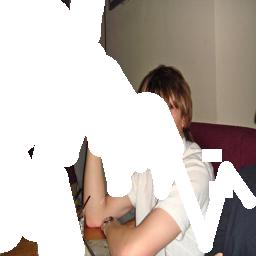}}&
	\frame{\includegraphics[width=\M\linewidth]{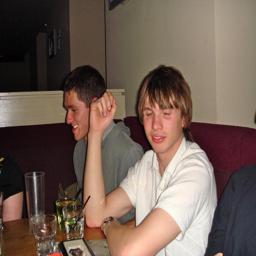}}&
	\frame{\includegraphics[width=\M\linewidth]{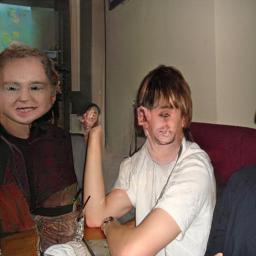}}&
	\frame{\includegraphics[width=\M\linewidth]{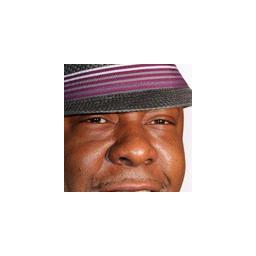}}&
	\frame{\includegraphics[width=\M\linewidth]{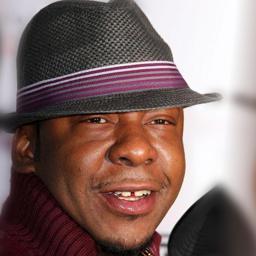}}&
	\frame{\includegraphics[width=\M\linewidth]{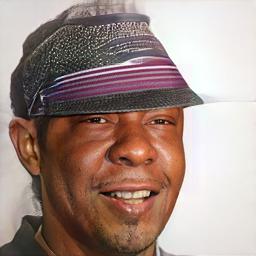}}  \\
	\scriptsize{Masked} & \scriptsize{Original} & \scriptsize{Ours} & \scriptsize{Masked} & \scriptsize{Original} & \scriptsize{Ours} 
\end{tabular}
\caption{Illustration of two failure cases of the proposed method.}
\label{fig:fail}
\end{figure}

\section{Conclusion}

In this paper, we propose an encoder-based GAN inversion method InvertFill for image inpainting. The encoder projects corrupted images into a latent space $\mathcal{F}\&\mathcal{W}^+$ with pre-modulation for learning more discriminative representation. The novel latent space $\mathcal{F}\&\mathcal{W}^+$ resolves the ``gapping'' issue when applied to GAN inversion in image inpainting. In addition, the soft-update mean latent dynamically samples diverse in-domain patterns, leading to more realistic textures. Extensive quantitative and qualitative comparisons demonstrate the superiority of our model over previous approaches and can cheaply support the semantically consistent completion of images or masks from unseen domains.  

\section*{Acknowledgment}

This work was supported by the Key Research Program of Frontier Sciences, CAS, Grant No. ZDBS-LY-JSC038. Libo Zhang was supported by the CAAI-Huawei MindSpore Open Fund and Youth Innovation Promotion Association, CAS (2020111). Heng Fan and his employer received no financial support for the research, authorship, and/or publication of this article.



\clearpage
%
%
\bibliographystyle{splncs04}
\bibliography{egbib}
\end{document}